\documentclass{article} %
\usepackage{iclr2027_conference,times}
\usepackage{graphicx}
\usepackage{amsmath,amsfonts,bm}

\usepackage[colorlinks=true,breaklinks=true,bookmarks=false,citecolor=green]{hyperref}
\usepackage{booktabs}
\usepackage{url}
\usepackage{xcolor}
\usepackage{multirow}
\usepackage{tabularx}
\usepackage{array}
\usepackage{algorithm}
\usepackage{algpseudocode}
\usepackage{amsmath,amssymb}
\newcommand{\dimmed}[1]{\textcolor{gray!40}{#1}}

\definecolor{rred}{RGB}{245, 152, 153}
\definecolor{oorange}{RGB}{253, 205, 154}

\title{MatAnyone-E: Sparse Memory Propagation for Efficient High-Resolution Video Matting}

\author{
\hspace{-2.5mm}\fontsize{9}{1}\selectfont
Shuyun Wang$^{1,2}$ \
Shangchen Zhou$^{2\dagger}$ \
Peiqing Yang$^{2}$ \
Yuekun Dai$^{2}$ \
Xin Yu$^{3\ast}$ \
Chen Change Loy$^{2\ast}$
\vspace{1mm}\\
\fontsize{9}{1}\selectfont
\hspace{-2mm}$^{1}$The University of Queensland  \hspace{0.1cm}
$^{2}$S-Lab, Nanyang Technological University  \hspace{0.1cm}
$^{3}$University of Adelaide
\vspace{1mm}\\
\hspace{-1.5mm}{\tt\small \url{https://shuyun-wang.github.io/projects/MatAnyone-E}}
}

\newcommand\blfootnote[1]{%
  \begingroup
  \renewcommand\thefootnote{}\footnote{#1}%
  \addtocounter{footnote}{-1}%
  \endgroup
}

\iclrfinalcopy %
\begin{document}

\maketitle
\fancyhead[L]{Preprint}

\begin{figure}[h]
\centering
\includegraphics[width=1\linewidth]{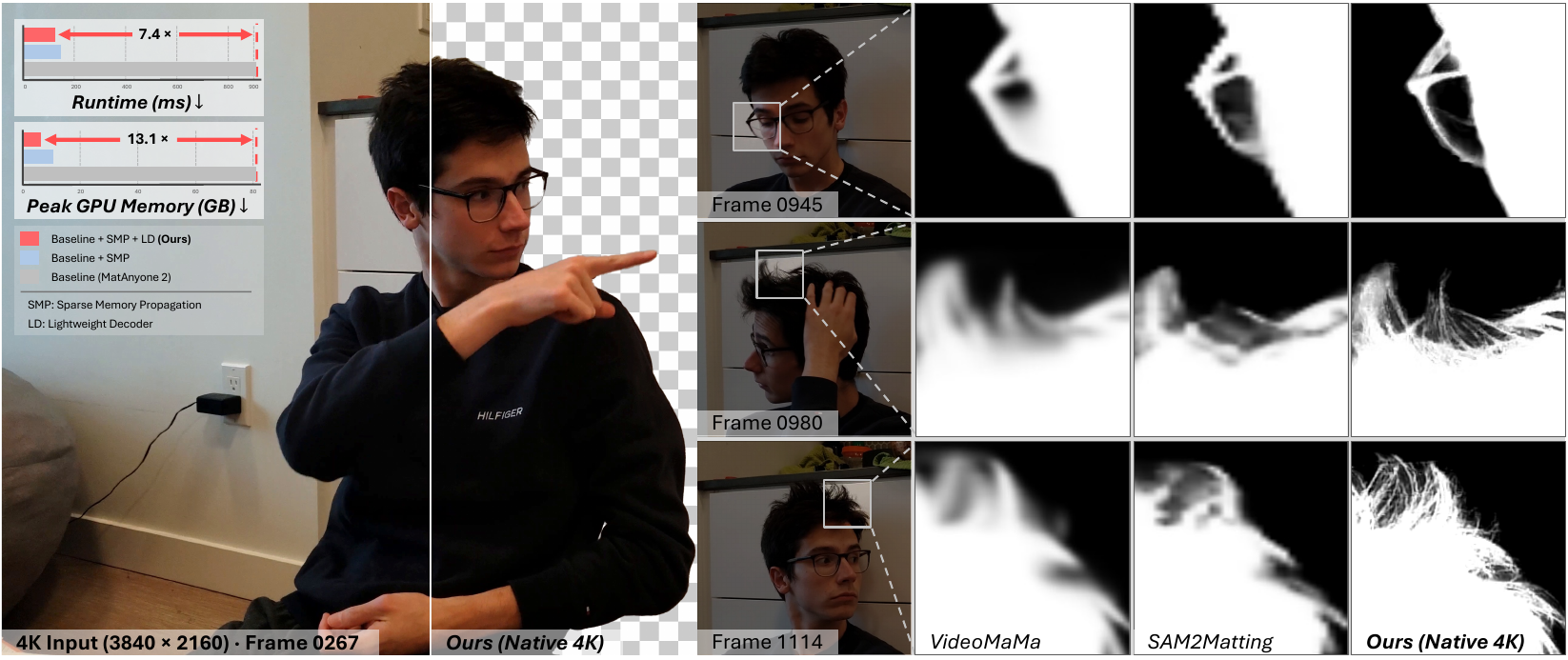}
\vspace{-4mm}
\caption{\textbf{Real-world 4K inference on an NVIDIA A800-SXM4-80GB.}
MatAnyone-E (Ours) achieves an estimated $7.4\times$ speedup and $13.1\times$ lower peak GPU memory usage than MatAnyone~2~\citep{matanyone2}, which runs out of memory.
By directly processing native 4K inputs, our method preserves finer details and boundaries compared to VideoMaMa~\citep{videomama} and SAM2Matting~\citep{sam2matting}.}
\label{fig:teaser}
\end{figure}

\begin{abstract}
High-resolution video matting requires preserving fine details, such as hair strands and soft transparency, while maintaining consistency across frames.
Memory-based methods achieve strong matting quality by matching current-frame features with information stored from earlier frames, but this process becomes increasingly costly as resolution grows.
We present MatAnyone-E, an efficient extension of MatAnyone~2 that focuses computation where it is most needed.
Our approach first performs low-resolution propagation to identify uncertain regions that require closer inspection and the memory candidates most relevant to them.
It then performs high-resolution matching only for these regions using the selected memory, while reusing low-resolution matching results elsewhere.
This strategy reduces both the number of locations requiring detailed matching and the amount of memory searched, while preserving the information needed to produce a complete matte.
We further reduce the cost of generating high-resolution mattes with a lightweight decoder designed around measured runtime bottlenecks and trained to reproduce the original decoder’s representations.
Experiments show that MatAnyone-E maintains matting quality comparable to MatAnyone~2 while achieving an estimated $7.4\times$ inference speedup and reducing peak GPU memory usage to $7.6\%$ of the usage of MatAnyone~2 at 4K.
With compilation-enabled inference on a single NVIDIA H200 GPU, it reaches 73.07 FPS at 1080p and 28.62 FPS at 4K.
\blfootnote{$^\dagger$Project lead.\quad  $^\ast$Corresponding authors.}
\end{abstract}

\section{Introduction}
Video matting estimates a continuous alpha matte for each frame, enabling foreground subjects to be composited onto new backgrounds.
High-resolution applications demand accurate recovery of hair strands, thin structures, and soft transparency, together with temporal consistency to prevent flickering.
Meeting these requirements, however, becomes computationally expensive as resolution increases, limiting the practical use of otherwise high-quality methods.

Existing video matting methods~\citep{oavm,late,attention,fast} address this challenge with different trade-offs.
Efficiency-oriented approaches, such as RVM~\citep{rvm} and SparseMat~\citep{sparsemat}, reduce computation through recurrent temporal modeling or selective high-resolution refinement, but can struggle to preserve fine structures and temporal consistency in challenging scenes.
Recent methods such as VideoMaMa~\citep{videomama} and SAM2Matting~\citep{sam2matting} leverage generative or segmentation priors, yet their restricted processing resolutions limit the detail they can recover from high-resolution inputs.
Memory-based methods, including MatAnyone~\citep{matanyone} and MatAnyone~2~\citep{matanyone2}, achieve strong matting quality by retrieving information from earlier frames.
However, matching current-frame features against stored memory and decoding high-resolution mattes incur substantial computational costs as resolution increases.
Although selective matte refinement has been explored, reducing the cost of the underlying temporal memory matching remains a distinct challenge.
This motivates our central question: how can we preserve the quality of memory-based video matting while avoiding exhaustive high-resolution matching and costly decoding?

The main computational bottleneck lies in comparing every current-frame feature with a large bank of features stored from earlier frames.
Increasing the video resolution enlarges both sets of features, causing the matching cost to grow rapidly even when the number of stored frames remains fixed.
Yet detailed matching is not equally important everywhere.
The model can rely more on information carried over from the previous frame at some locations, while others depend more strongly on newly retrieved information.
This suggests focusing matching computation on locations that require stronger updates from memory.
The difficulty is identifying these locations efficiently: the signal indicating how much each location needs an update is obtained only after the expensive matching has already been performed.

We address this dilemma with MatAnyone-E, an efficient extension of MatAnyone~2.
Our key idea is to use an inexpensive low-resolution propagation pass to guide subsequent computation.
This initial pass identifies locations that are likely to benefit most from closer inspection and the memory candidates most relevant to those locations.
We then perform matching at the native feature resolution only between these selected locations and memory candidates, reusing the upsampled low-resolution matching results elsewhere.
Selecting locations alone would still require searching the entire memory bank; selecting relevant memory as well reduces both sources of matching cost.
The resulting dense matching output is then used for downstream alpha prediction.

Beyond memory matching, producing the high-resolution alpha matte remains a substantial computational cost.
We therefore develop a lightweight decoder guided by measured runtime bottlenecks, since reducing arithmetic operations alone does not necessarily improve inference speed.
We train this decoder through knowledge distillation from the original decoder to preserve matting quality.
Together, selective memory matching and efficient decoding address the two main sources of high-resolution inference cost.

Experiments show that MatAnyone-E maintains matting quality comparable to MatAnyone~2 while substantially improving inference efficiency.
At 4K resolution, our complete model achieves an estimated $7.4\times$ inference speedup and reduces peak GPU memory usage to $7.6\%$ of MatAnyone~2. Compared with prior methods, MatAnyone-E provides a stronger balance between matting quality and inference efficiency at high resolutions.

\section{Related Work}
\label{sec:related_work}
\noindent \textbf{Video Matting.}
Video matting methods address the coupled challenges of alpha-detail recovery, temporal consistency, and computational efficiency.
RVM~\citep{rvm} uses recurrent feature aggregation for real-time human video matting.
SparseMat~\citep{sparsemat} exploits spatio-temporal sparsity for selective high-resolution refinement, while $\mu$Matting~\citep{uMatting} targets 4K matting with selective refinement and introduces the $\alpha$Matte4K dataset.
These efficiency-oriented methods~\citep{MODNet,bivm,bgmv2} highlight efficient temporal modeling and selective refinement as important strategies for high-resolution matting.
Another line of work~\citep{vmformer,adam,vim} leverages external guidance and pretrained priors.
FTP-VM~\citep{ftp} jointly models trimap propagation and matting.
MaGGIe~\citep{maggie} combines transformer attention and sparse convolution for mask-guided instance matting.
Matting Anything 2~\citep{mattinganything2} uses separate memory pathways for masks and trimaps.
SAM2Matting~\citep{sam2matting} connects video tracking~\citep{sam2,sam3} with dedicated matting heads.
GVM~\citep{gvm} exploits video diffusion priors~\citep{svd}, while VideoMaMa~\citep{videomama} leverages generative priors to convert segmentation masks into alpha mattes and train SAM2-Matte.

Closely related to our work, MatAnyone~\citep{matanyone} introduces region-adaptive memory fusion for stable target-assigned video matting.
MatAnyone~2~\citep{matanyone2} further improves matting through learned quality evaluation, large-scale real-world data curation, and reference-frame training.
We build on this memory-based framework and focus on reducing its high-resolution inference cost.
Whereas methods such as SparseMat and $\mu$Matting reduce computation through selective matte refinement, reducing the cost of the underlying temporal memory matching remains a distinct challenge.

\noindent \textbf{Efficient Video Object Segmentation.}
Memory-based video object segmentation (VOS) provides a foundation for temporal information propagation.
STCN~\citep{stcn} improves efficiency through shared frame correspondences and distance-based affinity.
XMem~\citep{xmem} combines sensory, working, and long-term memories to control memory growth over long videos.
Cutie~\citep{cuite} augments pixel-level memory reading with object-level queries for more robust target representations.
SAM~2~\citep{sam2} extends promptable segmentation to videos through streaming memory attention.
These methods highlight the importance of memory representation and efficient memory access for temporal propagation.

Recent work further reduces the cost of memory-based VOS.
EfficientTAM~\citep{efficienttam} combines a lightweight plain ViT encoder with spatially compressed memory attention.
EdgeTAM~\citep{edgetam} compresses frame memories using a 2D Spatial Perceiver with global and patch-level queries, supported by knowledge distillation.
Efficient-SAM2~\citep{efficientsam2} introduces object-aware window routing and sparse memory retrieval to reduce redundant computation.
These approaches demonstrate the effectiveness of memory compression and selective computation for efficient temporal propagation.
Our work extends this efficiency perspective to high-resolution video matting by using uncertainty estimated from inexpensive low-resolution propagation to determine where native-resolution matching is needed, together with query-dependent memory selection.

\section{Methodology}
\label{sec:method}

\label{sec:method_overview}
\begin{figure}[t]
\centering
\includegraphics[width=1\linewidth]{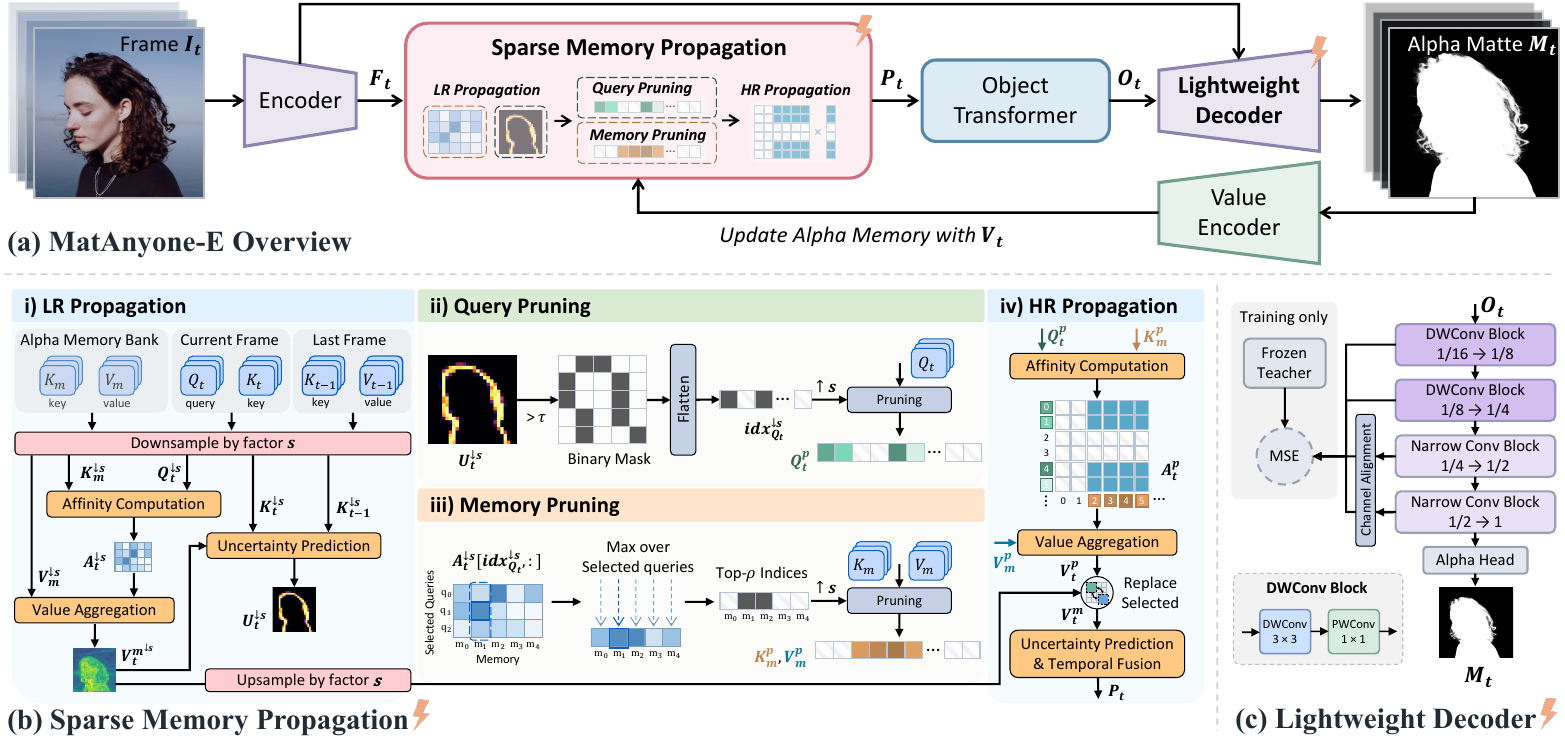}
\vspace{-3mm}
\caption{\textbf{Overview of MatAnyone-E.}
(a) MatAnyone-E retains the main architecture of MatAnyone~2~\citep{matanyone2}, while replacing its memory propagation and decoder with Sparse Memory Propagation and a lightweight decoder.
(b) Low-resolution (LR) propagation estimates uncertainty and query--memory affinities to select query locations and relevant memory for high-resolution (HR) matching.
HR propagation then performs matching only on the selected query--memory pairs, while reusing the upsampled LR readout elsewhere.
(c) The lightweight decoder adopts depthwise-separable blocks in its lower-resolution stages and narrower dense convolutions in its later decoder stages, guided by stage-wise latency profiling and trained with feature distillation from the MatAnyone~2 decoder.}
\label{fig:framework}
\vspace{-1mm}
\end{figure}

Given a video and a first-frame foreground mask, MatAnyone-E predicts an alpha matte $M_t$ for each frame $I_t$.
As illustrated in Fig.~\ref{fig:framework}(a), MatAnyone-E follows the main architecture of MatAnyone~2, but we modify two of the components whose computational cost grows most rapidly with resolution: memory propagation (Fig.~\ref{fig:framework}(b)) and decoding (Fig.~\ref{fig:framework}(c)).
Sparse Memory Propagation (Sec.~\ref{sec:sparse_memory_propagation}) uses low-resolution propagation to identify uncertain regions that require closer inspection and the memory candidates most relevant to them. It then performs high-resolution matching only for these regions using the selected memory, while reusing low-resolution matching results elsewhere.
A lightweight decoder (Sec.~\ref{sec:lightweight_decoder}) further reduces high-resolution decoding cost through profiling-guided design and feature distillation.

\subsection{Preliminaries}
Throughout this section, HR refers to the encoder's native memory-feature resolution, and LR refers to this feature grid spatially downsampled by a factor of $s$.

\noindent\textbf{Memory Matching.}
Let $Q_t\in\mathbb{R}^{N_q\times C_k}$ denote the current-frame queries, and let $K_m\in\mathbb{R}^{N_m\times C_k}$ and $V_m\in\mathbb{R}^{N_m\times C_v}$ denote the stored memory keys and values.
We omit batch and object dimensions for clarity.
MatAnyone~2 retrieves information from memory as:
\begin{equation}
\label{eq:memory_matching}
A_t=
\operatorname{Norm}_{m}\!\left(
\operatorname{Sim}(Q_t,K_m)
\right),
\qquad
V_t^m=A_tV_m ,
\end{equation}
where $A_t\in\mathbb{R}^{N_q\times N_m}$ contains query--memory affinities, $\operatorname{Sim}$ denotes the baseline distance-based similarity, and $\operatorname{Norm}_m$ retains the top-$k$ affinities of each query before softmax normalization over memory locations.
Although only a small number of memory locations contribute to each readout, finding these matches still requires comparing every query against the full memory bank.
Since both the query and memory grids grow with spatial resolution, such dense matching quickly becomes expensive at high resolution even when the number of stored frames is fixed.

\noindent\textbf{Temporal Fusion.}
After memory retrieval, MatAnyone~2 combines the retrieved value $V_t^m$ with the value $V_{t-1}$ propagated from the previous frame:
\begin{equation}
P_{t,i}
=
U_{t,i}\,V_{t,i}^{m}
+
(1-U_{t,i})\,V_{t-1,i},
\label{eq:temporal_fusion}
\end{equation}
where $U_{t,i}\in[0,1]$ is predicted from the retrieved memory readout $V_t^m$ by the uncertainty predictor.
This uncertainty does not measure prediction confidence; rather, it is associated with temporal alpha changes and acts as a per-location gate: small values favor information carried over from the previous frame in temporally stable regions, whereas large values place more weight on newly retrieved memory.
The uncertainty $U_t$ thus reflects how strongly each location relies on memory retrieval.

\subsection{Sparse Memory Propagation}
\label{sec:sparse_memory_propagation}
Although Eq.~\ref{eq:temporal_fusion} weights retrieved memory differently across locations, MatAnyone~2 still matches every query against the full memory bank.
Ideally, detailed matching should be reserved for locations that rely more heavily on newly retrieved information. However, identifying these locations is non-trivial because the fusion weights that indicate this reliance are available only after the memory readout has been computed.

The proposed Sparse Memory Propagation resolves this dependency with an inexpensive LR propagation pass. As shown in Fig.~\ref{fig:framework}(b),
the resulting coarse uncertainty $U_t^{\downarrow s}$ serves as a proxy for where HR matching is useful.
At the same time, the query--memory affinities already computed by this pass indicate which memory is relevant to those locations.
Hence, the same LR computation therefore answers two questions: where detailed matching should be performed, and which memory should participate in it. We detail the steps below:

\noindent\textbf{Low-Resolution Propagation.}
We first spatially downsample the queries, memory keys, and memory values by a factor of $s$ and apply the baseline memory matching:
\begin{equation}
\label{eq:lr_propagation}
A_t^{\downarrow s}
=
\operatorname{Norm}_{m}\!\left(
\operatorname{Sim}(Q_t^{\downarrow s},K_m^{\downarrow s})
\right),
\qquad
V_t^{m\downarrow s}
=
A_t^{\downarrow s}V_m^{\downarrow s}.
\end{equation}
Since both the query and memory grids contain approximately $1/s^2$ as many locations, this pass evaluates only about $1/s^4$ as many query--memory pairs as dense HR matching.
The LR readout $V_t^{m\downarrow s}$ is used to estimate uncertainty for query selection and is upsampled to provide a dense base for subsequent HR propagation, while the LR affinities $A_t^{\downarrow s}$ are reused to identify relevant memory.

\noindent\textbf{Query Pruning.}
We apply the uncertainty predictor inherited from MatAnyone~2 to the LR propagation features $K_t^{\downarrow s}$, $K_{t-1}^{\downarrow s}$, and $V_t^{m\downarrow s}$ to estimate $U_t^{\downarrow s}$.
We then select coarse locations whose uncertainty exceeds a threshold $\tau$:
\begin{equation}
\label{eq:query_pruning}
\mathrm{idx}_{Q_t}^{\downarrow s}
=
\bigl\{\, i \;\big|\; U_{t,i}^{\downarrow s}>\tau \,\bigr\}.
\end{equation}
Each selected LR location corresponds to an $s\times s$ region on the HR grid, and the HR queries within these regions form
$Q_t^p\in\mathbb{R}^{n_q\times C_k}$.
Detailed matching can therefore focus on a subset of query locations rather than the full HR grid.

\noindent\textbf{Memory Pruning.}
Query pruning alone is not sufficient, because each retained query would still be compared against the entire memory bank, whose size grows with both the resolution and the number of stored frames.
The LR pass has already established coarse correspondences between these queries and memory.
We reuse these affinities to identify memory locations that are relevant to the selected queries.
Specifically, for each LR memory location $j$, we compute:
\begin{equation}
\label{eq:memory_pruning}
w_j=
\max_{i\in\mathrm{idx}_{Q_t}^{\downarrow s}}
\bigl(A_t^{\downarrow s}\bigr)_{i,j}.
\end{equation}
The maximum assigns a high score to a memory location whenever it is strongly associated with any selected query.
We retain the highest-scoring fraction $\rho$ of memory locations and use their corresponding HR keys and values, denoted by $K_m^{p}$ and $V_m^{p}$, for subsequent HR matching.
This pruning is performed only for the current retrieval and does not modify the persistent memory bank.

\noindent\textbf{High-Resolution Propagation.}
After pruning both sides of the matching problem, HR matching is performed only between the selected queries and selected memory:
\begin{equation}
\label{eq:hr_propagation}
A_t^p=
\operatorname{Norm}_{m}\!\left(
\operatorname{Sim}(Q_t^p,K_m^{p})
\right),
\qquad
V_t^{p}=A_t^pV_m^{p}.
\end{equation}

We use the upsampled LR readout as a dense base and replace the selected locations with their refined HR values:
\begin{equation}
\label{eq:readout_reconstruction}
V_{t,i}^{m}
=
\begin{cases}
V_{t,i}^{p}, & i\text{ is selected},\\[2pt]
[\operatorname{Up}_s(V_t^{m\downarrow s})]_i, & \text{otherwise}.
\end{cases}
\end{equation}
where $\operatorname{Up}_s$ denotes spatial upsampling by a factor of $s$.
This yields a dense readout while restricting expensive high-resolution matching to the selected query--memory pairs.
Dense HR matching evaluates $N_qN_m$ query--memory pairs.
Sparse Memory Propagation first performs the LR pass, which evaluates approximately $N_qN_m/s^4$ pairs, and then performs HR matching only between the retained $n_q$ queries and $n_m$ memory locations, requiring $n_qn_m$ pairs.
Its total matching cost is therefore approximately $N_qN_m/s^4+n_qn_m$.
The dense readout is then processed by the uncertainty predictor and temporal fusion at the HR level, as in MatAnyone~2.

\subsection{Lightweight Decoder}
\label{sec:lightweight_decoder}
Beyond memory matching, high-resolution decoding remains a substantial source of inference cost.
We therefore develop a lightweight decoder guided by measured runtime behavior and trained with feature distillation from the MatAnyone~2 decoder.

\noindent\textbf{Profiling-Guided Decoder Design.}
As illustrated in Fig.~\ref{fig:framework}(c), both the original and lightweight decoders progressively upsample features through four stages, increasing the spatial resolution from $1/16$ to $1/8$, $1/4$, $1/2$, and finally the full input resolution.
While the original decoder uses dense convolutions throughout, simply replacing them with lower-FLOP depthwise-separable convolutions does not consistently reduce latency across stages.
In particular, at higher spatial resolutions, the reduction in arithmetic cost is increasingly offset by lower arithmetic intensity and increased memory-access overhead, so fewer FLOPs do not necessarily translate into faster execution.

We therefore adopt different lightweight designs at different stages.
For the $1/16\rightarrow1/8$ and $1/8\rightarrow1/4$ stages, we use a depthwise $3\times3$ convolution followed by a pointwise $1\times1$ convolution.
For the $1/4\rightarrow1/2$ and $1/2\rightarrow1$ stages, where reducing memory traffic becomes more important, we retain dense convolutions but reduce their channel widths.
This stage-wise design follows the measured runtime behavior of each stage rather than applying a single lightweight block throughout the decoder.
Detailed stage-wise profiling and analysis are provided in Appendix~\ref{app:decoder_analysis}.

\noindent\textbf{Feature Distillation.}
While these lightweight designs reduce inference cost, they might limit the decoder's representation capacity.
To prevent this, we apply feature distillation during training, so that the intermediate representations learned by the original decoder are preserved.

Specifically, we use the MatAnyone~2 decoder as a frozen teacher and the lightweight decoder as the student.
Given the same upstream representation, we extract intermediate features from all four corresponding decoder stages.
Let $F_l^{T}$ and $F_l^{S}$ denote the teacher and student features at stage $l$, respectively.
When their channel dimensions differ, a learnable $1\times1$ adapter projects the student feature to the teacher dimension; otherwise, the original student feature is used directly.
Let $\widetilde{F}_l^{S}$ denote the resulting aligned student feature.
We then minimize:
\begin{equation}
\label{eq:decoder_distillation}
\mathcal{L}_{\mathrm{distill}}
=
\sum_{l=1}^{4}
\left\|
\widetilde{F}_l^{S}-F_l^{T}
\right\|_2^2 .
\end{equation}

The distillation loss is optimized jointly with the MatAnyone~2 matting objectives as
$\mathcal{L}=\mathcal{L}_{\mathrm{mat}}+\lambda_{\mathrm{distill}}\mathcal{L}_{\mathrm{distill}}$,
where $\lambda_{\mathrm{distill}}$ controls the strength of feature distillation.

\section{Experiments}
\label{sec:experiments}
\begin{table*}[t]
\vspace{-3mm}
\centering
\caption{\textbf{Matting quality comparison at \(\boldsymbol{1920\times1080}\)}.
Best and second-best results are marked in \colorbox{rred}{\underline{red}} and \colorbox{oorange}{orange}, respectively.
Values in parentheses denote differences from MatAnyone~2.}
\vspace{1mm}
\renewcommand{\arraystretch}{1.15}
\renewcommand{\tabcolsep}{1.3mm}
\resizebox{\linewidth}{!}{%
\begin{tabular}{lccccccc|cc}
\toprule
Metric & RVM & FTP-VM & MaGGIe & GVM & MatAnyone & VideoMaMa & SAM2Matting & MatAnyone 2 & \textbf{Ours} \\
\midrule

\multicolumn{10}{@{}l@{}}{\textbf{VideoMatte HR}} \\ \midrule
MAD \(\downarrow\)   & 6.57 & 8.00 & 4.42 & 6.33 & 4.24 & 6.83 & 4.74 & \colorbox{rred}{\underline{4.10}} & \colorbox{oorange}{4.14}~(+0.04) \\
MSE \(\downarrow\)   & 1.93 & 3.24 & 0.40 & 2.08 & 0.33 & 2.75 & 0.51 & \colorbox{oorange}{0.28} & \colorbox{rred}{\underline{0.25}}~(-0.03) \\
Grad \(\downarrow\)  & 10.55 & 23.75 & 4.03 & 8.04 & 4.00 & 6.04 & 6.78 & \colorbox{oorange}{3.45} & \colorbox{rred}{\underline{3.22}}~(-0.23) \\
dtSSD \(\downarrow\) & 1.90 & 2.37 & 1.31 & 1.59 & 1.19 & 1.40 & 1.74 & \colorbox{rred}{\underline{1.15}} & \colorbox{oorange}{1.17}~(+0.02) \\
\midrule

\multicolumn{10}{@{}l@{}}{\textbf{YouTubeMatte HR}} \\ \midrule
MAD \(\downarrow\)   & 4.37 & 6.49 & 2.37 & 2.68 & 1.99 & 2.46 & 3.16 & \colorbox{rred}{\underline{1.61}} & \colorbox{oorange}{1.62}~(+0.01) \\
MSE \(\downarrow\)   & 2.25 & 4.58 & 0.98 & 1.34 & 0.71 & 1.28 & 1.50 & \colorbox{oorange}{0.50} & \colorbox{rred}{\underline{0.45}}~(-0.05) \\
Grad \(\downarrow\)  & 15.10 & 29.78 & 7.69 & 8.38 & 8.91 & \colorbox{rred}{\underline{4.68}} & 13.07 & 7.14 & \colorbox{oorange}{6.13}~(-1.01) \\
dtSSD \(\downarrow\) & 2.28 & 2.41 & 1.77 & 1.72 & 1.65 & \colorbox{rred}{\underline{1.35}} & 2.10 & 1.53 & \colorbox{oorange}{1.48}~(-0.05) \\
\midrule

\multicolumn{10}{@{}l@{}}{\textbf{CRGNN}} \\ \midrule
MAD \(\downarrow\)   & 5.98 & 6.64 & 9.50 & 5.03 & 5.76 & 5.18 & 5.80 & \colorbox{rred}{\underline{4.24}} & \colorbox{oorange}{4.28}~(+0.04) \\
MSE \(\downarrow\)   & 2.79 & 3.54 & 6.11 & 2.15 & 3.04 & 2.13 & 2.66 & \colorbox{oorange}{2.00} & \colorbox{rred}{\underline{1.85}}~(-0.15) \\
Grad \(\downarrow\)  & 13.68 & 15.10 & 16.51 & 14.28 & 15.55 & 13.78 & 16.28 & \colorbox{rred}{\underline{11.74}} & \colorbox{oorange}{12.87}~(+1.13) \\
dtSSD \(\downarrow\) & 5.36 & 5.98 & 6.02 & 4.86 & 5.44 & 4.82 & 5.42 & \colorbox{rred}{\underline{4.54}} & \colorbox{oorange}{4.61}~(+0.07) \\

\bottomrule
\end{tabular}}
\vspace{-2mm}
\label{tab:quality_efficiency}
\end{table*}

\noindent\textbf{Training and Inference Details.}
We initialize MatAnyone-E from the pretrained MatAnyone~2 model and train it on 8 NVIDIA A800-80GB GPUs with a total batch size of 16.
Each training sample consists of 8 consecutive frames randomly cropped to $480\times480$, and training uses a combination of existing video matting datasets and the VMReal training set~\cite{matanyone2}.
Feature distillation is applied at all four decoder stages with $\lambda_{\mathrm{distill}}=0.001$.
The feature widths across the lightweight decoder hierarchy are \([256,128,64,32,16]\).
Feature adapters are used only where the student and teacher channel dimensions differ and are removed after training.
For Sparse Memory Propagation, we use a downsampling factor of $s=2$, an uncertainty threshold of $\tau=0.15$, and retain the top $15\%$ of memory candidates ($\rho=0.15$).
Unless otherwise specified, all experiments are conducted without model compilation.

\noindent\textbf{Evaluation Settings.}
We evaluate matting quality on the high-resolution subsets of VideoMatte~\cite{bgmv2} and YouTubeMatte~\cite{matanyone}, as well as the CRGNN benchmark~\cite{rvm}, at $1920\times1080$.
We report mean absolute difference (MAD), mean squared error (MSE), gradient error (Grad), and temporal error (dtSSD), where lower values indicate better performance.
For CRGNN, inference is performed over the complete video sequence, while quantitative evaluation is conducted only on annotated frames.
Unless otherwise specified, efficiency is measured on an NVIDIA A800-SXM4-80GB GPU.
For MatAnyone-E, runtime is measured at steady state after the memory bank reaches its configured capacity.

\begin{table}[t]
\centering
\vspace{-4mm}
\caption{\textbf{Inference efficiency across input resolutions.}
Measured on an NVIDIA A800-SXM4-80GB; FLOPs are reported per frame in TFLOPs and memory as peak allocated GPU memory in GB.
Values in parentheses for ``Ours'' denote speedup or reduction relative to MatAnyone~2, and ``(C.)'' denotes compilation-enabled inference.
Grayed-out results are not directly comparable because the corresponding methods operate at fixed processing resolutions rather than the native input resolution.
Asterisks indicate MatAnyone~2 4K results extrapolated from shorter-memory measurements because inference with a full memory bank exceeds available GPU memory; ``-'' indicates an out-of-memory result.}
\vspace{1mm}
\renewcommand{\arraystretch}{1.15}
\renewcommand{\tabcolsep}{1.8mm}
\resizebox{\linewidth}{!}{%
\begin{tabular}{lcccc|cc|c}
\toprule
 & MaGGIe & GVM & VideoMaMa & SAM2Matting & MatAnyone 2 & Ours & Ours (C.) \\
\midrule
Params (M) & 30.91 & 1621.02 & 2335.29 & 478.53 & 35.25 & 31.32 & 31.32 \\
\cmidrule(lr){1-8}

\multicolumn{8}{l}{\textbf{1024\(\times\)1024}} \\ \cmidrule(lr){1-8}
FPS \(\uparrow\) & 22.8 & 1.8 & 2.4 & 14.1 & 29.5 & 41.3~($\times$1.40) & 47.3 \\
FLOPs (\(\times10^{12}\)) \(\downarrow\) & 0.32 & 24.04 & 16.48 & 7.22 & 0.54 & 0.34~($\times$1.59) & 0.34 \\
Mem.\ (GB) \(\downarrow\) & 0.80 & 20.45 & 15.14 & 3.99 & 1.60 & 0.92~($\times$1.74) & 0.61 \\
\cmidrule(lr){1-8}

\multicolumn{8}{l}{\textbf{1920\(\times\)1080}} \\ \cmidrule(lr){1-8}
FPS \(\uparrow\) & \dimmed{19.4} & 0.84 & \dimmed{2.5} & \dimmed{14.0} & 13.6 & 26.6~($\times$1.96) & 37.6 \\
FLOPs (\(\times10^{12}\)) \(\downarrow\) & \dimmed{0.56} & 53.95 & \dimmed{16.48} & \dimmed{7.22} & 1.20 & 0.69~($\times$1.74) & 0.69 \\
Mem.\ (GB) \(\downarrow\) & \dimmed{1.19} & 33.99 & \dimmed{15.14} & \dimmed{4.01} & 5.47 & 1.66~($\times$3.30) & 1.07 \\
\cmidrule(lr){1-8}

\multicolumn{8}{l}{\textbf{2560\(\times\)1440}} \\ \cmidrule(lr){1-8}
FPS \(\uparrow\) & \dimmed{18.4} & 0.43 & \dimmed{2.5} & \dimmed{13.8} & 5.6 & 16.5~($\times$2.95) & 25.1 \\
FLOPs (\(\times10^{12}\)) \(\downarrow\) & \dimmed{0.56} & 111.61 & \dimmed{16.48} & \dimmed{7.22} & 2.45 & 1.24~($\times$1.98) & 1.24 \\
Mem.\ (GB) \(\downarrow\) & \dimmed{1.21} & 55.47 & \dimmed{15.10} & \dimmed{4.03} & 16.20 & 3.04~($\times$5.33) & 2.02 \\
\cmidrule(lr){1-8}

\multicolumn{8}{l}{\textbf{3840\(\times\)2160}} \\ \cmidrule(lr){1-8}
FPS \(\uparrow\) & \dimmed{19.3} & - & \dimmed{2.4} & \dimmed{13.3} & 1.1$^{*}$ & 8.1~($\times$7.36) & 13.6 \\
FLOPs (\(\times10^{12}\)) \(\downarrow\) & \dimmed{0.56} & - & \dimmed{16.48} & \dimmed{7.22} & 7.76$^{*}$ & 2.72~($\times$2.85) & 2.72 \\
Mem.\ (GB) \(\downarrow\) & \dimmed{1.26} & - & \dimmed{15.14} & \dimmed{4.08} & 79.69$^{*}$ & 6.08~($\times$13.11) & 3.89 \\
\bottomrule
\end{tabular}}
\label{tab:efficiency}
\vspace{-4mm}
\end{table}

\subsection{Quantitative Evaluations}
\label{sec:exp_quantitative}

\noindent\textbf{Comparison with State-of-the-art Methods.}
We compare MatAnyone-E with RVM~\citep{rvm}, FTP-VM~\citep{ftp}, MaGGIe~\citep{maggie}, GVM~\citep{gvm}, MatAnyone~\citep{matanyone}, VideoMaMa~\citep{videomama}, SAM2Matting~\citep{sam2matting}, and its baseline, MatAnyone~2~\citep{matanyone2}.
Table~\ref{tab:quality_efficiency} reports results on three video matting benchmarks.
Compared with MatAnyone~2, MatAnyone-E largely preserves matting quality: MAD differs by at most $0.04$ and dtSSD by at most $0.07$ across all three datasets.
It also achieves lower MSE on VideoMatte ($0.25$ vs.\ $0.28$), YouTubeMatte ($0.45$ vs.\ $0.50$), and CRGNN ($1.85$ vs.\ $2.00$).

Compared with recent methods, MatAnyone-E outperforms SAM2Matting across all reported metrics and datasets, and achieves lower MAD and MSE than VideoMaMa across all three benchmarks.
For example, on YouTubeMatte, MatAnyone-E achieves an MAD of $1.62$ vs.\ $3.16$ and an MSE of $0.45$ vs.\ $1.50$ compared with SAM2Matting.
Overall, these results show that MatAnyone-E maintains the strong matting quality of MatAnyone~2.

\noindent\textbf{High-Resolution Inference.}
Table~\ref{tab:efficiency} compares inference efficiency across four input resolutions.
At 4K, MatAnyone-E reaches $8.1$ FPS with $6.08$ GB peak memory.
Since MatAnyone~2 exceeds available GPU memory with a full memory bank at 4K, its runtime and memory are extrapolated from shorter-memory measurements, with details provided in Appendix~\ref{app:baseline_profiling}.
This corresponds to an estimated $7.4\times$ speedup and $13.1\times$ memory reduction over MatAnyone~2.
MatAnyone-E also achieves $2.0\times$ and $2.9\times$ speedups at 1080p and 1440p, respectively.
Compilation further improves throughput to $13.6$ FPS at 4K.
MatAnyone-E maintains high throughput and low peak memory as input resolution increases.

\subsection{Qualitative Evaluations}
\label{sec:exp_qualitative}
\begin{figure*}[t]
\centering
\vspace{-1mm}
\includegraphics[width=\linewidth]{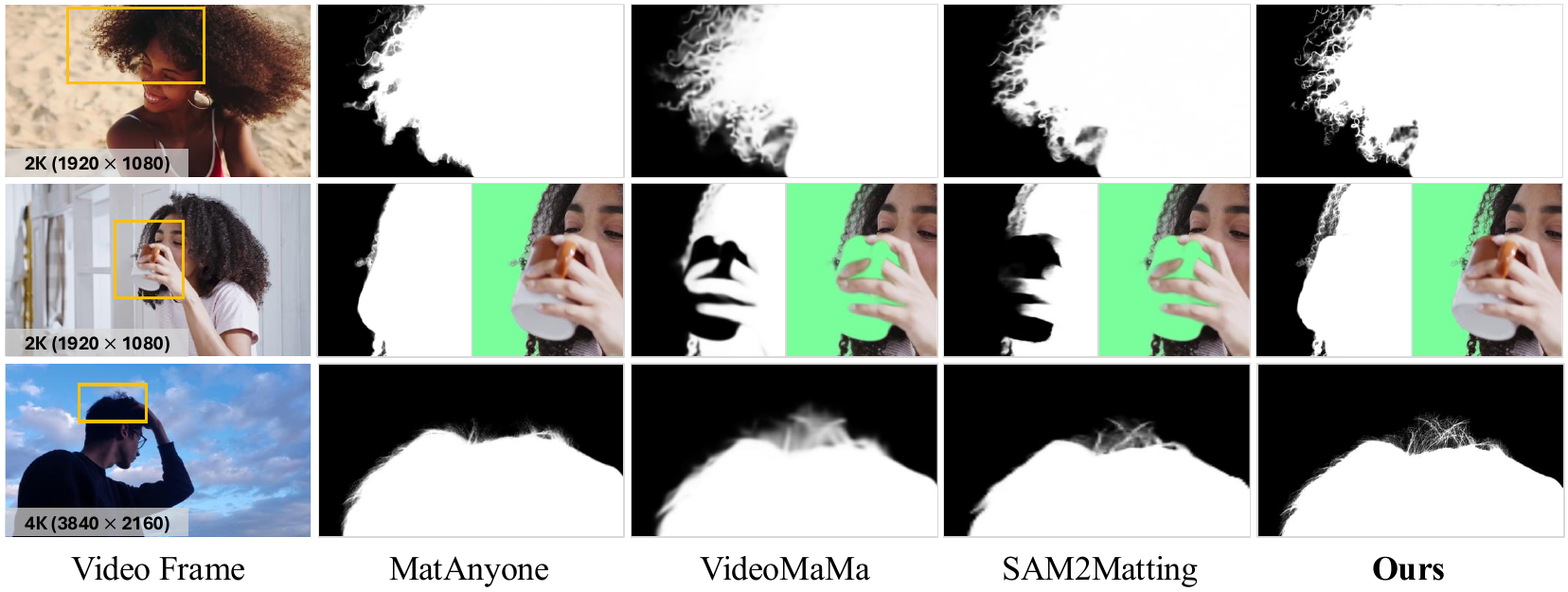}
\vspace{-5mm}
\caption{\textbf{Qualitative comparison.}
Examples illustrate fine-detail preservation, semantic correctness in challenging foreground regions, and high-resolution detail recovery.
MatAnyone-E better preserves thin hair structures and local boundaries, particularly in the 4K example.}
\label{fig:11}
\end{figure*}

\begin{figure*}[t]
\centering
\includegraphics[width=\linewidth]{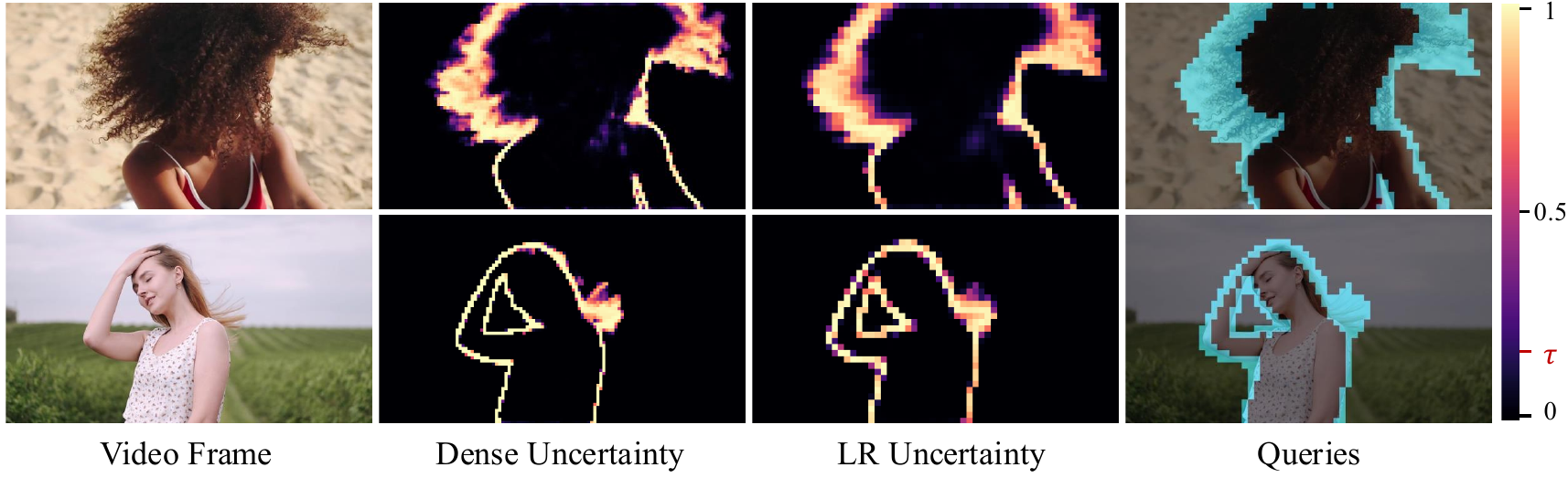}
\vspace{-5mm}
\caption{\textbf{Query selection on CRGNN.}
Low-resolution uncertainty follows the main spatial distribution of dense uncertainty and selects queries around object boundaries, fine structures, and other uncertain regions for high-resolution matching. Please zoom in to inspect fine details.}
\label{fig:12}
\vspace{-1mm}
\end{figure*}

\noindent\textbf{Qualitative Comparison.}
Figure~\ref{fig:11} compares MatAnyone-E with MatAnyone, VideoMaMa, and SAM2Matting on representative challenging examples.
In regions containing fine hair and soft boundaries, MatAnyone-E recovers cleaner and more complete alpha structures with fewer missing or blurred details.
It also better preserves challenging foreground structures, such as hands and objects interacting with the subject, while maintaining accurate local boundaries.
The advantage becomes more evident at 4K resolution, where MatAnyone-E processes the native-resolution input and preserves finer hair structures than methods operating at restricted processing resolutions.

\noindent\textbf{Query-Selection Visualization.}
Figure~\ref{fig:12} compares uncertainty estimated from dense propagation with that obtained from low-resolution propagation, together with the query locations selected for high-resolution matching.
The low-resolution estimate retains the main spatial distribution of dense uncertainty.
This supports its use as a low-cost proxy for identifying query locations that benefit from high-resolution matching.

\subsection{Ablation Studies}
\label{sec:exp_ablation}
\begin{table}[t]
\vspace{-2mm}
\centering
\caption{\textbf{Component ablation on CRGNN at $1920\times1080$.}
SMP denotes Sparse Memory Propagation, and LD denotes the lightweight decoder.
The baseline is the original MatAnyone~2 model; combining SMP and LD gives the full MatAnyone-E model.
Memory is reported as peak allocated GPU memory in GB.}
\label{tab:main_ablation}
\vspace{1mm}
\renewcommand{\arraystretch}{1.15}
\renewcommand{\tabcolsep}{1.8mm}
\resizebox{\linewidth}{!}{%
\begin{tabular}{l|cccc|ccc}
\toprule
Configuration & MAD $\downarrow$ & MSE $\downarrow$ & Grad $\downarrow$ & dtSSD $\downarrow$
& Latency (ms) $\downarrow$ & FPS $\uparrow$ & Peak Mem. (GB) $\downarrow$ \\
\midrule
Baseline          & 4.24 & 2.00 & 11.74 & 4.54 & 73.53 & 13.6 & 5.47 \\
+ SMP             & 4.26 & 1.94 & 13.00 & 4.69 & 44.18 & 22.6 & 2.78 \\
+ LD             & 4.27 & 1.84 & 12.83 & 4.59 & 66.61 & 15.0 & 4.35 \\
+ SMP + LD (Ours) & 4.28 & 1.85 & 12.87 & 4.61 & 37.59 & 26.6 & 1.66 \\
\bottomrule
\end{tabular}%
}
\vspace{0mm}
\end{table}

\begin{table}[t]
  \centering
    \caption{\textbf{Comparison of HR query-selection strategies on CRGNN at $1920\times1080$.}
    All selective strategies use approximately the same HR query budget.
    The baseline is the MatAnyone~2 with dense HR matching, while ``LR only'' uses only the LR readout without further HR matching.
    Lower is better for all quality metrics and latency.}
  \label{tab:query_selection_ablation}
  \renewcommand{\arraystretch}{1.15}
  \renewcommand{\tabcolsep}{2.5mm}
  \vspace{1mm}
  \resizebox{\linewidth}{!}{%
    \begin{tabular}{lcccccc}
      \toprule
      Method & MAD $\downarrow$ & MSE $\downarrow$ & Grad $\downarrow$ & dtSSD $\downarrow$ & Latency (ms) $\downarrow$ & HR Queries \\
      \midrule
      Baseline & 4.24 & 2.00 & 11.74 & 4.54 & 73.53 & 100.00\% \\
      LR only                & 4.35 & 2.00 & 13.23 & 4.80 & 41.85 & 0.00\% \\
      Uniform                & 4.39 & 2.05 & 13.26 & 4.82 & 43.82 & 6.98\% \\
      Alpha boundary         & 4.34 & 2.00 & 13.20 & 4.79 & 43.46 & 6.99\% \\
      \midrule
      Uncertainty-guided (Ours) & 4.26 & 1.94 & 13.02 & 4.70 & 42.92 & 6.99\% \\
      \bottomrule
    \end{tabular}%
  }
  \vspace{-2mm}
\end{table}
\noindent\textbf{Component Ablation.}
As shown in Table~\ref{tab:main_ablation}, both Sparse Memory Propagation (SMP) and the lightweight decoder (LD) improve inference efficiency while maintaining comparable matting quality.
Compared with the MatAnyone~2 baseline, SMP reduces latency from 73.53 ms to 44.18 ms and peak GPU memory from 5.47 GB to 2.78 GB, while LD reduces them to 66.61 ms and 4.35 GB, respectively.
Combining both components gives the full MatAnyone-E model, achieving 37.59 ms latency, 26.6 FPS, and 1.66 GB peak memory.

\noindent\textbf{Query-Selection Strategy.}
As shown in Table~\ref{tab:query_selection_ablation}, we compare different strategies for selecting query locations for HR matching using the MatAnyone~2 checkpoint.
The dense baseline performs HR matching at all query locations, while ``LR only'' uses only the LR readout without further HR matching.
Under a similar HR query budget, uncertainty-guided selection achieves better matting quality than uniform sampling and alpha-boundary selection, while maintaining comparable latency.
This supports using the uncertainty estimated from LR propagation to identify locations that benefit most from HR matching.

\section{Conclusion}
\label{sec:conclusion}

We presented MatAnyone-E, an efficient framework for high-resolution video matting.
Our key insight is that dense high-resolution memory matching is largely redundant: only a subset of query locations requires precise retrieval, and only a subset of memory is relevant to them.
By restricting native-resolution matching to selected query--memory pairs and reusing the upsampled low-resolution readout elsewhere, Sparse Memory Propagation substantially reduces the cost of temporal memory retrieval.
A lightweight decoder further reduces high-resolution decoding cost through profiling-guided design and feature distillation.
MatAnyone-E preserves the strong matting quality of MatAnyone~2 while achieving an estimated $7.4\times$ inference speedup and $13.1\times$ lower peak GPU memory at 4K.
We believe selective high-resolution computation offers a practical direction for scaling memory-based video matting to high resolutions.

\clearpage

\bibliography{iclr2027_conference}
\bibliographystyle{iclr2027_conference}

\clearpage
\appendix
\section*{Appendix}

The appendix provides additional implementation details, efficiency analyses, and qualitative results that complement the main paper:

\begin{itemize}
    \item Appendix~\ref{app:implementation}: Multi-level Sparse Memory Propagation
    \item Appendix~\ref{app:efficiency_analysis}: Additional Efficiency Analysis
    \item Appendix~\ref{app:memory_pruning}: Analysis of Memory Pruning
    \item Appendix~\ref{app:decoder_analysis}: Analysis of the Lightweight Decoder
    \item Appendix~\ref{app:cross_gpu}: Efficiency Across GPU Architectures
    \item Appendix~\ref{app:qualitative}: Additional Qualitative Results
\end{itemize}

\section{Multi-level Sparse Memory Propagation}
\label{app:implementation}
\begin{algorithm}[ht]
\caption{Multi-level Sparse Memory Propagation.
\(X^{\downarrow s}\): \(s{\times}\) spatial downsampling (\(X^{\downarrow 1}{=}X\));
\(\operatorname{Up}_r\): \(r{\times}\) spatial upsampling;
\(\operatorname{Unc}\): the inherited uncertainty predictor;
\(c(i)\): the \(s_0\)-grid location containing \(i\);
\(X[\mathcal{I}]\): entries of \(X\) inside the regions covered by \(\mathcal{I}\).
With a single level (\(L{=}1\), \(s_0{=}s\)), it reduces to the formulation in the main paper.}
\label{alg:multilevel_smp}
\begin{algorithmic}[1]
\Require queries \(Q_t\) (\(H{\times}W\) grid); memory \(K_m,V_m\);
spatial budget \(M\); uncertainty threshold \(\tau\); keep ratio \(\rho\)
\Ensure memory readout \(V_t^m\)

\Function{Levels}{$H,W,M$}
    \State \(s \gets 2\)
    \While{\(\lceil H/s\rceil\lceil W/s\rceil > M\)}
        \State \(s \gets 2s\)
    \EndWhile
    \State \Return \((s,s/2,\ldots,2,1)\)
\EndFunction

\State \((s_0,\ldots,s_L)\gets\Call{Levels}{H,W,M}\)

\Statex \textit{Low-resolution propagation}
\State \(A_t^{\downarrow s_0}
\gets
\operatorname{Norm}_m(
\operatorname{Sim}(Q_t^{\downarrow s_0},
K_m^{\downarrow s_0}))\)

\State \(V_t^{m\downarrow s_0}
\gets
A_t^{\downarrow s_0}V_m^{\downarrow s_0}\)

\For{\(\ell=1,\ldots,L\)}
    \State \(U_t^{\downarrow s_{\ell-1}}
    \gets
    \operatorname{Unc}(V_t^{m\downarrow s_{\ell-1}})\)

    \Statex \hspace{\algorithmicindent}\textit{Query pruning}
    \State \(\mathcal{I}
    \gets
    \{i\mid U_{t,i}^{\downarrow s_{\ell-1}}>\tau\}\)
    \State \(Q_t^p\gets Q_t^{\downarrow s_\ell}[\mathcal{I}]\)

    \Statex \hspace{\algorithmicindent}\textit{Memory pruning}
    \State \(w_j
    \gets
    \max_{i\in\mathcal{I}}
    (A_t^{\downarrow s_0})_{c(i),j}\)
    \State \(\mathcal{J}
    \gets
    \text{top-}\rho\text{ fraction of }\{w_j\}\)
    \State \(K_m^p\gets K_m^{\downarrow s_\ell}[\mathcal{J}]\);
           \(V_m^p\gets V_m^{\downarrow s_\ell}[\mathcal{J}]\)

    \Statex \hspace{\algorithmicindent}\textit{Sparse propagation}
    \State \(A_t^p
    \gets
    \operatorname{Norm}_m(
    \operatorname{Sim}(Q_t^p,K_m^p))\)
    \State \(V_t^p\gets A_t^pV_m^p\)

    \State \(V_t^{m\downarrow s_\ell}
    \gets
    \operatorname{Up}_{s_{\ell-1}/s_\ell}
    (V_t^{m\downarrow s_{\ell-1}})\)

    \State \(V_t^{m\downarrow s_\ell}[\mathcal{I}]
    \gets
    V_t^p\)
\EndFor

\State \Return \(V_t^m=V_t^{m\downarrow s_L}\)
\end{algorithmic}
\end{algorithm}

For high-resolution inputs, we extend Sparse Memory Propagation (SMP) to multiple refinement levels.
The number of levels is determined by the spatial size of the memory feature grid.
Starting from the native feature resolution \(H\times W\), we repeatedly downsample by a factor of two until the coarsest grid contains no more than \(M\) spatial locations.
This produces a sequence of scales
\((s_0,s_1,\ldots,s_L)\), where \(s_L=1\) corresponds to the native memory-feature resolution.

Propagation first operates densely at the coarsest level.
The resulting memory readout is used to estimate uncertainty, and only query locations whose uncertainty exceeds the threshold are propagated at the next finer level.
For these selected queries, we further restrict matching to a subset of memory candidates identified from the coarse affinities.
Specifically, memory candidates are ranked according to their maximum affinity to the selected query regions, and only the top \(\rho\) fraction is retained.
Sparse propagation is then performed between the selected queries and retained memory candidates.
The refined readouts replace the corresponding regions of the upsampled coarse result, while the remaining locations directly retain the coarse prediction.
This process is repeated until the native memory-feature resolution is reached.

Algorithm~\ref{alg:multilevel_smp} summarizes the complete procedure.
When only a single refinement level is required, the algorithm reduces to the formulation described in the main paper.
For all 4K experiments reported in the main paper, we consistently use a
\(4\!\rightarrow\!2\!\rightarrow\!1\) propagation schedule with the spatial
budget set to \(M=4096\).

\section{Additional Efficiency Analysis}
\label{app:efficiency_analysis}

\subsection{Runtime Bottlenecks of MatAnyone~2}
\label{app:baseline_profiling}
\begin{table*}[ht]
\centering
\caption{Per-module runtime breakdown of MatAnyone~2 across resolutions and memory-bank sizes. Runtime is reported in milliseconds per frame. The 4K result with five memory frames is estimated because the full configuration exceeds GPU memory.}
\vspace{1mm}
\label{tab:runtime_breakdown}
\renewcommand{\arraystretch}{1.2}
\setlength{\tabcolsep}{5pt}
\resizebox{\linewidth}{!}{%
\begin{tabular}{lrrrrr}
\toprule
Module & 1080p (mem=5) & 1440p (mem=5) & 4K (mem=2) & 4K (mem=3) & 4K (mem=5, est.) \\
\midrule
Encoder            & 9.12  & 15.81  & 34.11  & 34.10  & 34.11 \\
Memory Propagation & 34.35 & 115.94 & 314.58 & 469.65 & 779.80 \\
Object Transformer & 7.56  & 8.33   & 13.86  & 13.91  & 14.01 \\
Decoder            & 17.59 & 30.37  & 67.21  & 67.22  & 67.22 \\
Value Encoder      & 4.54  & 7.53   & 16.22  & 16.22  & 16.22 \\
Others             & 0.73  & 0.85   & 1.25   & 1.34   & 1.51 \\
\midrule
Total latency      & 73.88 & 178.83 & 445.10 & 600.30 & $\approx$911.00 \\
\bottomrule
\end{tabular}}
\end{table*}

To better understand the source of high-resolution inference cost, we profile MatAnyone~2 at the module level across input resolutions and memory-bank sizes.
As shown in Table~\ref{tab:runtime_breakdown}, memory propagation increasingly dominates runtime as the spatial resolution grows, accounting for \(46\%\) of the total runtime at 1080p and \(65\%\) at 1440p.
At 4K, the bottleneck becomes substantially more pronounced: increasing the memory bank from two to five frames raises the memory-propagation cost from \(314.58\) ms to an estimated \(779.80\) ms, while the runtime of the remaining modules remains nearly unchanged.
Consequently, the total per-frame latency increases from \(445.1\) ms to approximately \(911\) ms.
The decoder is the second-largest high-resolution component, but unlike memory propagation, its cost is largely independent of the memory-bank size.
These results identify dense memory propagation as the primary runtime bottleneck of MatAnyone~2 at high resolution.

\subsection{Component-wise Efficiency Analysis}
\label{app:component_efficiency}
\begin{table*}[ht]
  \centering
  \caption{Inference efficiency of MatAnyone-E components across resolutions.
  QP and MP denote query pruning and memory pruning, respectively, and LD denotes the lightweight decoder.
  Each row cumulatively adds the indicated component to the previous configuration.
  Latency is reported in milliseconds, peak memory in GB, and computational cost in TFLOPs per frame.
  Speedup is relative to MatAnyone~2 at the same resolution.}
  \vspace{1mm}
  \label{tab:efficiency_components}
  \renewcommand{\arraystretch}{1.2}
  \setlength{\tabcolsep}{5pt}
  \resizebox{\linewidth}{!}{%
    \begin{tabular}{l r r r r r r}
      \toprule
      Method & Latency (ms) $\downarrow$ & FPS $\uparrow$ & Speedup $\uparrow$ &
      Peak Mem. (GB) $\downarrow$ & FLOPs (TFLOPs) $\downarrow$ & Params (M) \\
      \midrule
      \multicolumn{7}{@{}l@{}}{\textbf{1080p}} \\
      \midrule
      MatAnyone 2          & 73.53 & 13.6 & -- & 5.47 & 1.20 & 35.25 \\
      + QP                 & 42.90 & 23.3 & 1.71$\times$ & 2.69 & 0.97 & 35.25 \\
      + MP                 & 44.18 & 22.6 & 1.66$\times$ & 2.78 & 0.96 & 35.25 \\
      + LD                 & 37.59 & 26.6 & 1.96$\times$ & 1.66 & 0.69 & 31.32 \\
      + Compilation        & 26.60 & 37.6 & 2.76$\times$ & 1.07 & 0.69 & 31.32 \\
      \midrule
      \multicolumn{7}{@{}l@{}}{\textbf{1440p}} \\
      \midrule
      MatAnyone 2          & 178.57 & 5.6 & -- & 16.20 & 2.45 & 35.25 \\
      + QP                 & 71.81 & 13.9 & 2.49$\times$ & 4.60 & 1.74 & 35.25 \\
      + MP                 & 69.70 & 14.3 & 2.56$\times$ & 4.84 & 1.71 & 35.25 \\
      + LD                 & 60.61 & 16.5 & 2.95$\times$ & 3.04 & 1.24 & 31.32 \\
      + Compilation        & 39.84 & 25.1 & 4.48$\times$ & 2.02 & 1.24 & 31.32 \\
      \midrule
      \multicolumn{7}{@{}l@{}}{\textbf{2160p}} \\
      \midrule
      MatAnyone 2          & 909.09 & 1.1 & -- & 79.69 & 7.76 & 35.25 \\
      + QP                 & 157.67 & 6.3 & 5.77$\times$ & 10.10 & 3.91 & 35.25 \\
      + MP                 & 143.55 & 7.0 & 6.33$\times$ & 10.18 & 3.77 & 35.25 \\
      + LD                 & 123.46 & 8.1 & 7.36$\times$ & 6.08 & 2.72 & 31.32 \\
      + Compilation        & 73.53 & 13.6 & 12.36$\times$ & 3.89 & 2.72 & 31.32 \\
      \midrule
      \multicolumn{7}{@{}l@{}}{\textbf{4320p}} \\
      \midrule
      MatAnyone 2          & \multicolumn{2}{c}{OOM} & -- & $>$80 & -- & 35.25 \\
      + QP                 & 1174.04 & 0.9 & -- & 45.79 & 17.27 & 35.25 \\
      + MP                 & 578.00 & 1.7 & -- & 39.87 & 15.29 & 35.25 \\
      + LD                 & 488.37 & 2.0 & -- & 23.79 & 11.05 & 31.32 \\
      + Compilation        & 300.74 & 3.3 & -- & 14.87 & 11.05 & 31.32 \\
      \bottomrule
    \end{tabular}%
  }
\end{table*}

We next isolate the contribution of each component to inference efficiency.
Table~\ref{tab:efficiency_components} progressively introduces query pruning (QP), memory pruning (MP), the lightweight decoder (LD), and compilation on top of MatAnyone~2.

Query pruning provides the largest reduction in memory-propagation cost, with increasingly pronounced gains at higher resolutions.
At 4K, QP reduces latency from \(909.09\) ms to \(157.67\) ms and peak memory from \(79.69\) GB to \(10.10\) GB.
Memory pruning further restricts matching along the memory dimension.
Its effect is limited at 1080p and 1440p, but becomes substantially more important at larger spatial scales.
At 8K, adding MP on top of QP reduces latency from \(1174.04\) ms to \(578.00\) ms and peak memory from \(45.79\) GB to \(39.87\) GB.

The lightweight decoder provides a complementary reduction in dense high-resolution computation.
At 4K, it reduces latency from \(143.55\) ms to \(123.46\) ms, peak memory from \(10.18\) GB to \(6.08\) GB, and computational cost from \(3.77\) to \(2.72\) TFLOPs per frame.
Compilation further improves runtime without changing the reported FLOPs or model parameters, reducing the 4K latency from \(123.46\) ms to \(73.53\) ms.

Overall, the results show that the main efficiency gains of MatAnyone-E come from reducing dense memory matching, while memory pruning becomes increasingly beneficial at very high resolutions and the lightweight decoder provides complementary savings in dense decoding.

\subsection{Resolution Dependence of Sparse Memory Propagation}
\label{app:smp_resolution}
Sparse Memory Propagation relies on low-resolution uncertainty and affinities to guide query and memory selection, and its effectiveness therefore depends on the spatial resolution of the underlying feature grid.
At lower input resolutions, the downsampled feature map can become too coarse to reliably localize uncertain regions or distinguish relevant memory candidates, limiting the benefit of selective matching.

This effect is particularly noticeable for query pruning.
As shown in Table~\ref{tab:efficiency_components}, its efficiency gain becomes more pronounced as the input resolution increases: the benefit is relatively modest at 1080p, becomes clearer at 1440p, and grows substantially at 4K and 8K.
Memory pruning shows a similar resolution-dependent trend.
Its additional gain is limited at 1080p and 1440p, but becomes much more significant at 8K, where the memory dimension is considerably larger.

These results indicate that Sparse Memory Propagation is most beneficial at high spatial resolutions, while its advantage can be limited when the low-resolution feature grid becomes too small.

\section{Analysis of Memory Pruning}
\label{app:memory_pruning}

\begin{table}[ht]
\centering
\caption{Recall of dense memory matching under different memory pruning ratios.}
\label{tab:memory_pruning_recall}
\vspace{1mm}
\renewcommand{\arraystretch}{1.15}
\setlength{\tabcolsep}{6pt}
\begin{tabular}{lcccc}
\toprule
Keep ratio & KV retained & Recall count & Recall weight & Top-1 hit \\
\midrule
5\%  & 4.94\%  & 0.8318 & 0.8333 & 0.9363 \\
10\% & 9.84\%  & 0.9503 & 0.9508 & 0.9879 \\
15\% & 14.76\% & 0.9805 & 0.9807 & 0.9962 \\
20\% & 19.71\% & 0.9908 & 0.9909 & 0.9985 \\
30\% & 29.67\% & 0.9977 & 0.9977 & 0.9997 \\
50\% & 49.70\% & 0.9999 & 0.9999 & 1.0000 \\
\bottomrule
\end{tabular}
\end{table}

We further analyze how well the pruned memory set preserves the candidates selected by dense matching.
Table~\ref{tab:memory_pruning_recall} reports the retained KV ratio together with three recall statistics under different pruning ratios.
Here, recall count measures the fraction of dense-matching candidates retained after pruning, recall weight measures the fraction of their affinity mass retained, and Top-1 hit measures the fraction of queries whose highest-affinity memory candidate is preserved.

The relevant memory candidates are highly concentrated.
Retaining only \(10\%\) of the memory bank already recovers approximately \(95\%\) of the dense-matching candidates and weights, while preserving the original top-1 match for \(98.8\%\) of queries.
At our default keep ratio of \(15\%\), the retained memory set preserves \(98.1\%\) of both candidate count and affinity weight, with a top-1 hit rate of \(99.6\%\).
Increasing the keep ratio beyond \(15\%\) yields diminishing improvements in recall.

These results indicate that coarse affinities provide a reliable signal for identifying query-relevant memory candidates.
We therefore use a \(15\%\) keep ratio in our experiments as a trade-off between retaining relevant memory and pruning redundant memory before high-resolution matching.

\section{Analysis of the Lightweight Decoder}
\label{app:decoder_analysis}

\begin{table}[ht]
\centering
\caption{Stage-wise profiling of the MatAnyone~2 decoder.
AI denotes arithmetic intensity, and ``Regime'' indicates whether each block is compute-bound or memory-bound.}
\vspace{1mm}
\label{tab:stage_profiling}
\renewcommand{\arraystretch}{1.1}
\setlength{\tabcolsep}{8pt}
\resizebox{0.78\linewidth}{!}{%
\begin{tabular}{lccccc}
\toprule
Block & Params (K) & Precision & Ridge Point & AI & Regime \\
\midrule
{up\_16\_8}       & 475.5  & FP16 TC   & 153.0 & 530.0 & Compute \\
{up\_8\_4}        & 295.2  & FP16 TC   & 153.0 & 461.0 & Compute \\
{up\_4\_2}        & 119.0  & FP16 TC   & 153.0 & 265.0 & Compute \\
\midrule
{up\_2\_1}        & 12.6   & FP16 TC   & 153.0 & 71.0  & Memory \\
{pred\_mat}       & 0.1    & FP32/TF32 & 76.5  & 4.2   & Memory \\
\bottomrule
\end{tabular}
}
\end{table}

We further profile the decoder stages to characterize their different computational regimes.
Table~\ref{tab:stage_profiling} reports the parameter count, arithmetic intensity, and hardware regime of each decoder block.
The lower-resolution stages exhibit relatively high arithmetic intensity and remain compute-bound, whereas the final high-resolution stages have substantially lower arithmetic intensity and become memory-bound.

This motivates us to adopt a stage-dependent lightweight design rather than a uniform operator replacement across all decoder stages.
For the compute-bound lower-resolution stages, we use depthwise-separable convolutions to reduce arithmetic cost.
For the high-resolution stages, where memory traffic becomes increasingly important, we instead reduce channel width while retaining dense convolutions.
Together, these choices reduce decoder cost while accounting for the different computational characteristics of individual stages.

\section{Efficiency Across GPU Architectures}
\label{app:cross_gpu}

\begin{table*}[t]
\centering
\vspace{-3mm}
\caption{Inference efficiency of MatAnyone-E across GPU architectures.
Each GPU reports FPS / peak memory (GB). ``w/o'' and ``w/'' denote inference without and with compilation.}
\vspace{1mm}
\label{tab:efficiency_cross_gpu}
\renewcommand{\arraystretch}{1.18}
\setlength{\tabcolsep}{6pt}
\resizebox{\linewidth}{!}{%
\begin{tabular}{cc cc cc cc cc cc}
\toprule
& &
\multicolumn{2}{c}{RTX 3090} &
\multicolumn{2}{c}{RTX 5090} &
\multicolumn{2}{c}{RTX A6000} &
\multicolumn{2}{c}{A800} &
\multicolumn{2}{c}{H200 NVL} \\
\cmidrule(lr){3-4}\cmidrule(lr){5-6}\cmidrule(lr){7-8}
\cmidrule(lr){9-10}\cmidrule(lr){11-12}
Resolution & Comp. & FPS & Mem. & FPS & Mem. & FPS & Mem. & FPS & Mem. & FPS & Mem. \\
\midrule
\multirow{2}{*}{$1024^2$}
& w/o & 33.60 & 0.87 & 70.38 & 0.89 & 34.63 & 0.87 & 41.37 & 0.92 & 71.57 & 0.89 \\
& w/  & 45.97 & 0.50 & 95.53 & 0.52 & 50.97 & 0.50 & 47.37 & 0.61 & 102.10 & 0.52 \\
\midrule
\multirow{2}{*}{1080p}
& w/o & 19.81 & 1.61 & 39.91 & 1.71 & 19.29 & 1.61 & 26.55 & 1.66 & 45.22 & 1.63 \\
& w/  & 29.37 & 0.87 & 61.53 & 1.11 & 30.44 & 0.87 & 37.64 & 1.07 & 73.07 & 1.05 \\
\midrule
\multirow{2}{*}{1440p}
& w/o & 11.99 & 2.98 & 22.42 & 3.00 & 11.48 & 2.98 & 16.51 & 3.04 & 29.22 & 3.00 \\
& w/  & 18.27 & 1.89 & 36.10 & 1.90 & 18.43 & 1.89 & 25.11 & 2.02 & 52.89 & 1.90 \\
\midrule
\multirow{2}{*}{2160p}
& w/o & 5.69 & 6.09 & 10.54 & 6.11 & 5.51 & 6.09 & 8.14 & 6.08 & 14.38 & 6.12 \\
& w/  & 9.00 & 3.16 & 19.10 & 3.18 & 9.40 & 3.16 & 13.63 & 3.89 & 28.62 & 3.18 \\
\bottomrule
\end{tabular}}
\end{table*}

To evaluate the efficiency of MatAnyone-E across hardware platforms, we benchmark it on five GPUs spanning consumer, professional, and datacenter classes.
Table~\ref{tab:efficiency_cross_gpu} reports throughput and peak GPU memory across four resolutions, with and without compilation.

Across all tested GPUs, MatAnyone-E exhibits consistent scaling trends as the input resolution increases.
Compilation further improves throughput and reduces peak memory across devices.
At 4K, the compiled model reaches \(9.0\) FPS on an RTX 3090, \(19.1\) FPS on an RTX 5090, \(9.4\) FPS on an RTX A6000, \(13.6\) FPS on an A800-SXM4-80GB, and \(28.6\) FPS on an H200 NVL.
These results show that the efficiency characteristics of MatAnyone-E remain consistent across a wide range of GPU architectures.

\section{Additional Qualitative Results}
\label{app:qualitative}

\begin{figure*}[t]
\centering
\includegraphics[width=\linewidth]{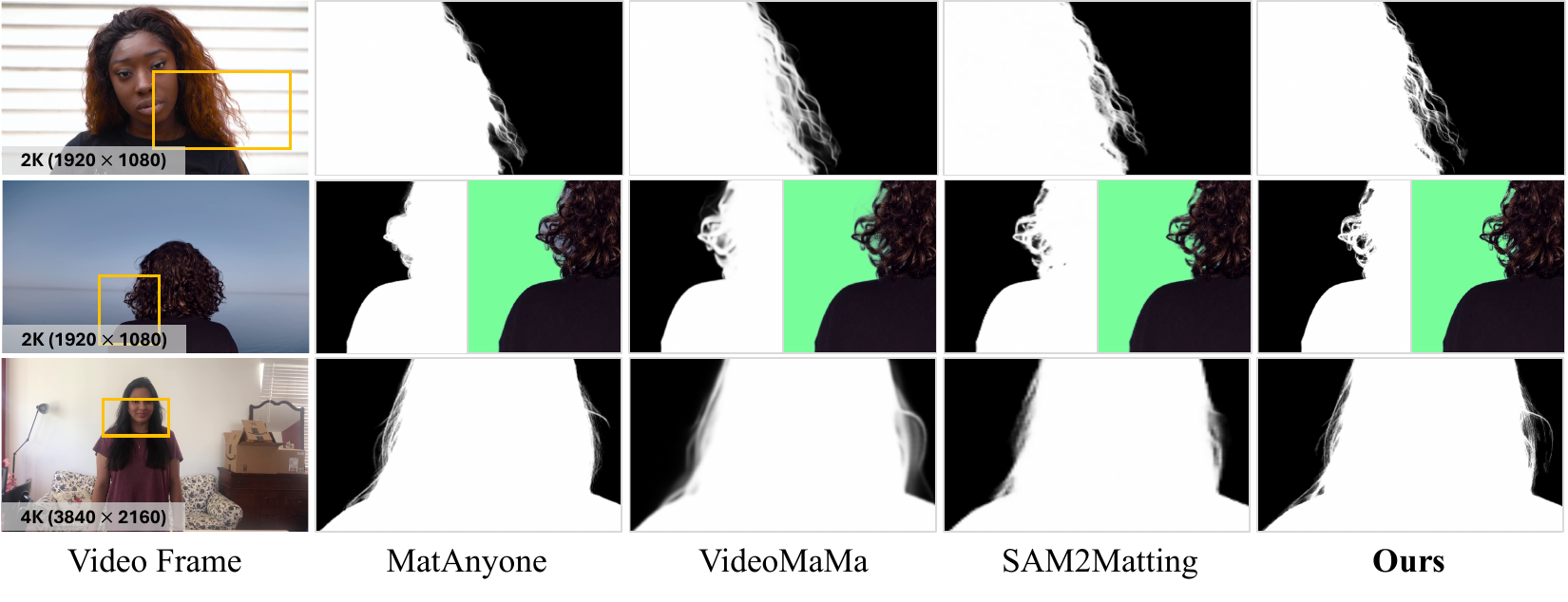}
\vspace{-6mm}
\caption{\textbf{Additional qualitative comparisons at 2K and 4K resolutions.}
We show additional examples comparing MatAnyone-E with MatAnyone, VideoMaMa, and SAM2Matting.}
\label{fig:app_qualitative}
\vspace{-2mm}
\end{figure*}

\noindent\textbf{Additional Qualitative Comparison.}
Figure~\ref{fig:app_qualitative} provides additional qualitative comparisons at 2K and 4K resolutions, complementing the examples shown in the main paper.

\begin{figure*}[t]
\centering
\includegraphics[width=\linewidth]{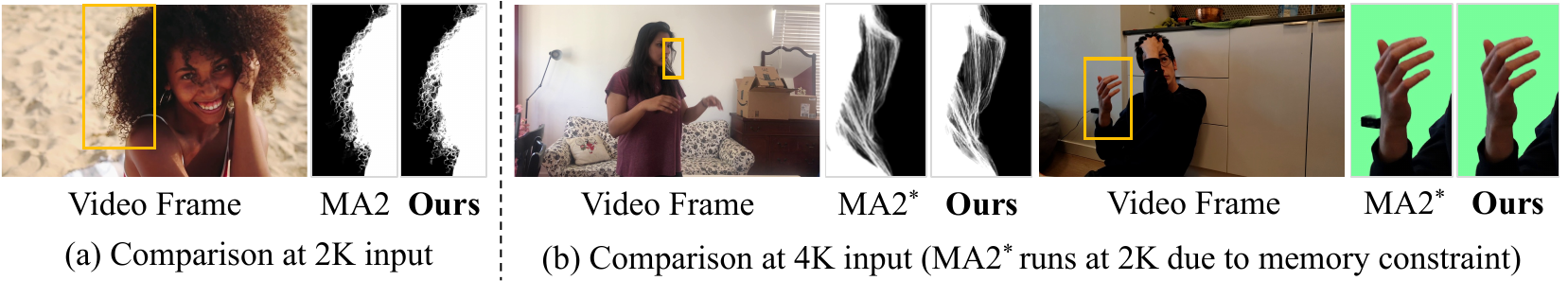}
\vspace{-6mm}
\caption{\textbf{Comparison with MatAnyone~2 at high resolution.}
MA2 denotes MatAnyone~2.
(a) Comparison at 2K input resolution, where MatAnyone-E achieves detail accuracy comparable to that of MA2.
(b) Comparison at 4K input resolution, where MA2$^{*}$ is evaluated at 2K because its full 4K configuration exceeds available GPU memory, while MatAnyone-E directly processes the native 4K input.
As a result, MA2$^{*}$ gives blurry details due to direct upsampling, and sometimes makes semantic mistakes when differentiating foreground from background.}
\label{fig:app_ma2_highres}
\end{figure*}

\noindent\textbf{Comparison with MatAnyone~2.}
Figure~\ref{fig:app_ma2_highres} provides additional comparisons with the direct baseline.
At 2K, MatAnyone-E preserves finer hair structures, while the advantage becomes more apparent for 4K inputs, where MatAnyone-E can directly process the native resolution.

\begin{figure*}[ht]
\centering
\includegraphics[width=\linewidth]{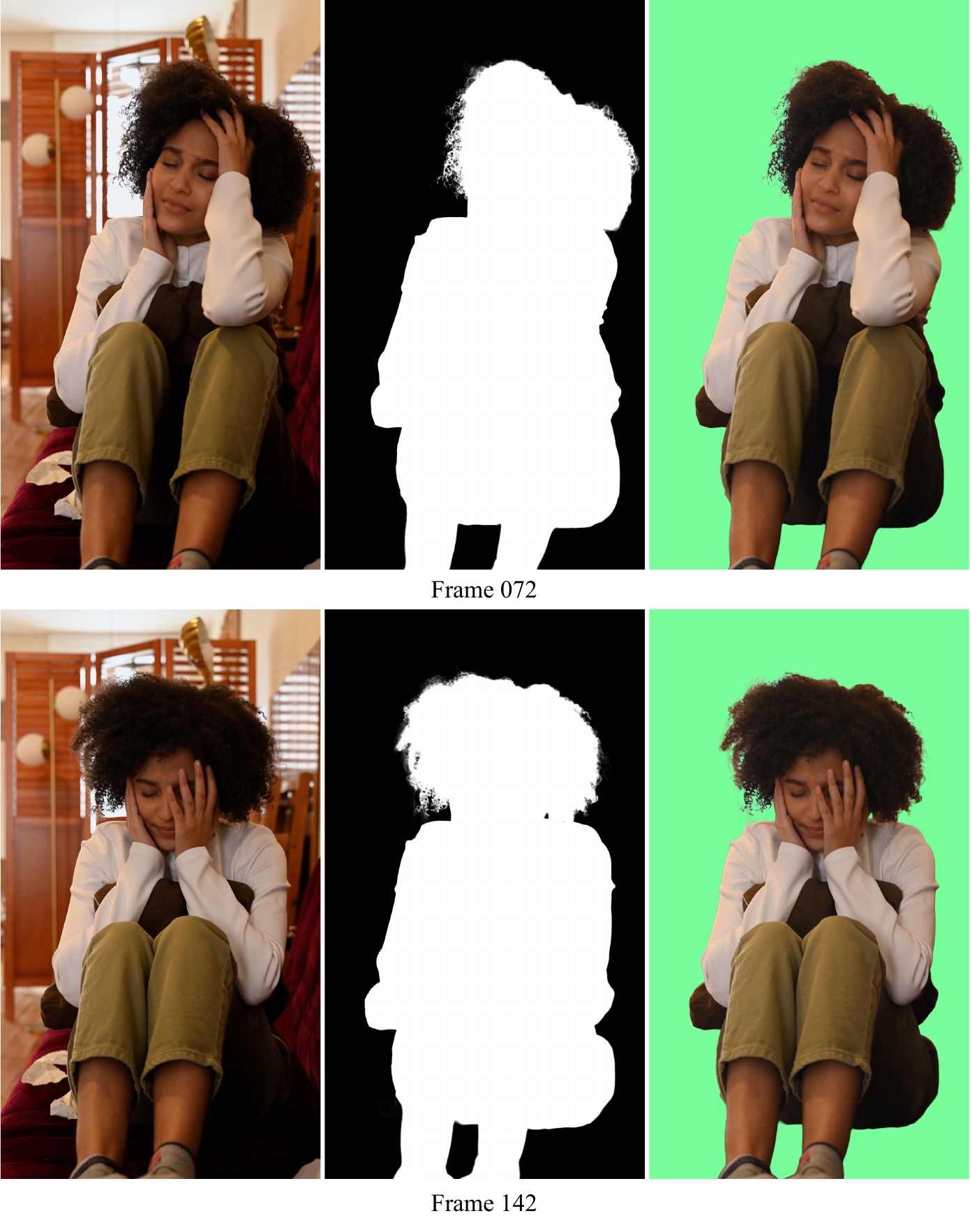}
\vspace{-8mm}
\caption{\textbf{Additional 4K matting results on a challenging real-world sequence.}
For each example, we show the input frame, predicted alpha matte, and the foreground composited on a green background.
MatAnyone-E preserves fine and complex hair structures across frames, including thin strands and highly irregular boundaries around the curly hair.}
\label{fig:app_amateur_green}
\end{figure*}

\begin{figure*}[ht]
\centering
\includegraphics[width=\linewidth]{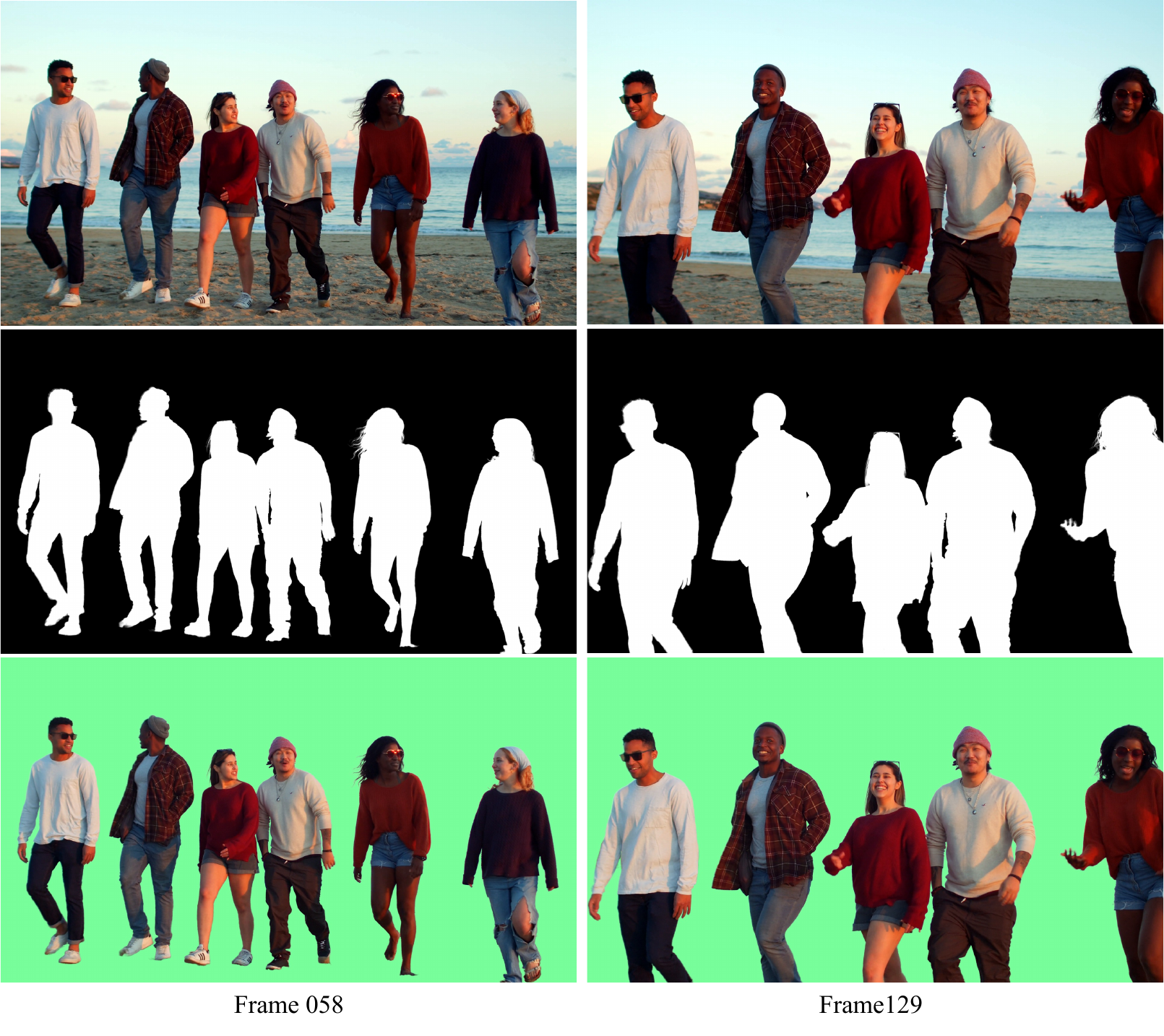}
\vspace{-8mm}
\caption{\textbf{4K matting results on a multi-person sequence (3840$\times$2160).}
We show two frames with the input images (top), predicted alpha mattes (middle), and foregrounds composited on a green background (bottom).}
\label{fig:app_hr_frames}
\end{figure*}

\begin{figure*}[ht]
\centering
\includegraphics[width=\linewidth]{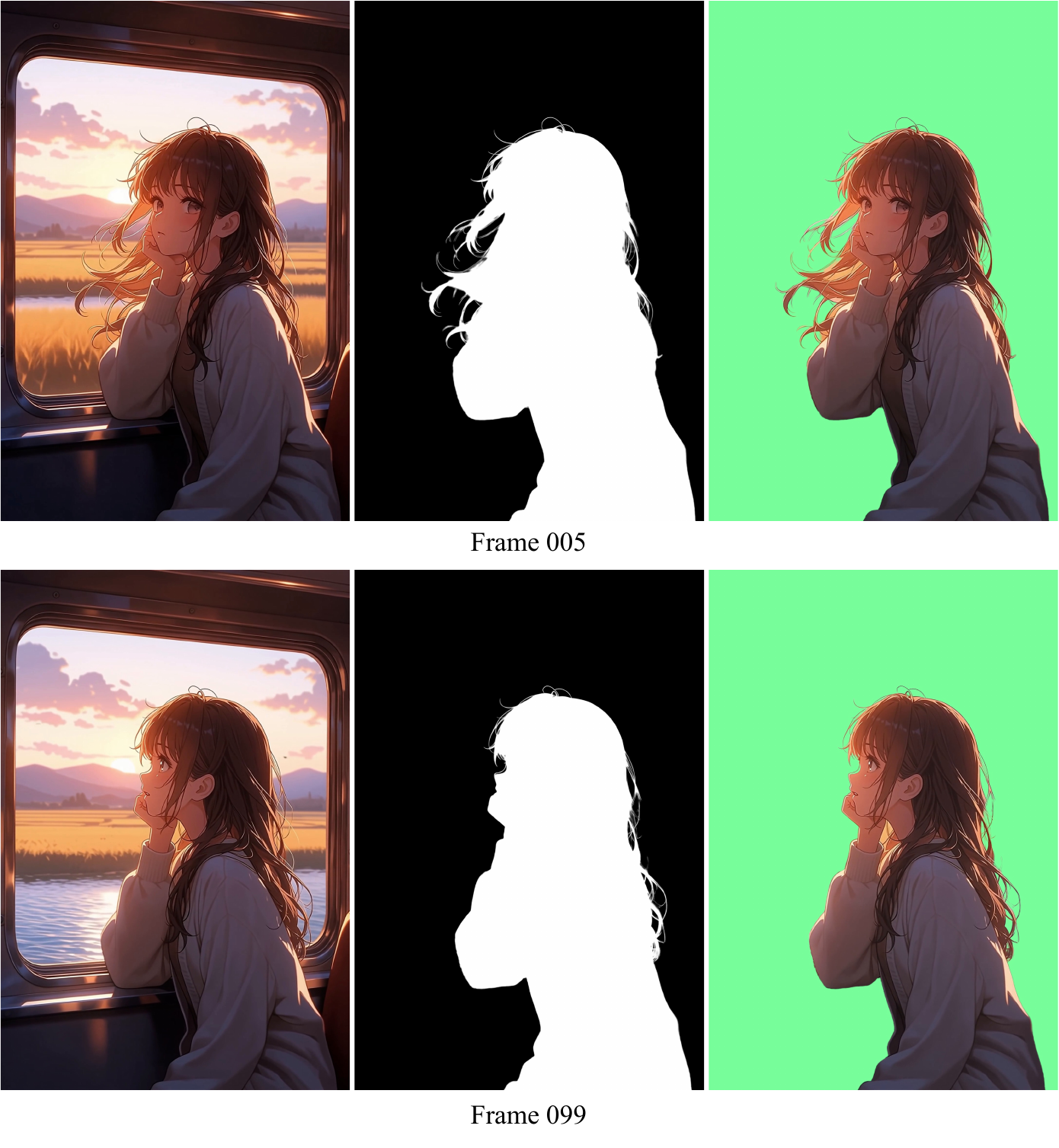}
\vspace{-8mm}
\caption{\textbf{High-resolution matting results on an AI-generated sequence (1440$\times$2144).}
We show two frames with the input images (left), predicted alpha mattes (middle), and foregrounds composited on a green background (right).}
\label{fig:app_hr_overall}
\end{figure*}

\noindent\textbf{Additional High-Resolution Results.}
Figure~\ref{fig:app_amateur_green} presents qualitative results on a real-world 4K sequence with complex curly hair.
Despite the dense and irregular hair boundaries, MatAnyone-E preserves fine foreground structures across frames.
We further demonstrate MatAnyone-E on three additional high-resolution sequences covering diverse scenarios: multi-person scenes (Fig.~\ref{fig:app_hr_frames}) and AI-generated content (Fig.~\ref{fig:app_hr_overall}).
These results illustrate the ability of MatAnyone-E to maintain detailed foreground boundaries and temporally consistent matting results across different types of high-resolution content.

\end{document}